\documentclass[letterpaper, 10 pt, conference]{ieeeconf}  
\usepackage{bm}
\usepackage{amsmath}
\usepackage{amssymb}
\usepackage{graphicx}
\usepackage{multirow}
\usepackage[table,xcdraw]{xcolor}
\usepackage{arydshln} 
\usepackage[normalem]{ulem}
\useunder{\uline}{\ul}{}
\usepackage{color}
\usepackage{colortbl}  
\usepackage{multirow} 
\usepackage[ruled]{algorithm2e}
\newtheorem{remark}{\textbf{Remark}}

\usepackage{arydshln} 

\usepackage{booktabs}
\usepackage{stfloats}
\usepackage{multirow}
\usepackage{hyperref}
\hypersetup{
	colorlinks=true,
 	citecolor=blue,
	linkcolor=red,
	filecolor=blue,      
	urlcolor=blue
}

\IEEEoverridecommandlockouts                              

\usepackage[space]{cite}

\title{\LARGE \bf
Consistent Distributed Cooperative Localization: A Coordinate Transformation Approach
}

\newcommand{\bmp}{\boldsymbol{\rm{ p }}}

\newcommand{\bmu}{\boldsymbol{\rm{ u }}}
\newcommand{\bmv}{\boldsymbol{\rm{ v }}}

\newcommand{\bmx}{\boldsymbol{\rm{ x }}}
\newcommand{\bmy}{\boldsymbol{\rm{ y }}}
\newcommand{\bmz}{\boldsymbol{\rm{ z }}}

\newcommand{\bmF}{\boldsymbol{\rm{F}}}
\newcommand{\bmG}{\boldsymbol{\rm{G}}}
\newcommand{\bmH}{\boldsymbol{\rm{H}}}
\newcommand{\bmI}{\boldsymbol{\rm{I}}}
\newcommand{\bmJ}{\boldsymbol{\rm{J}}}

\newcommand{\bmM}{\boldsymbol{\rm{M}}}
\newcommand{\bmN}{\boldsymbol{\rm{N}}}

\newcommand{\bmP}{\boldsymbol{\rm{P}}}
\newcommand{\bmQ}{\boldsymbol{\rm{Q}}}
\newcommand{\bmR}{\boldsymbol{\rm{R}}}

\newcommand{\bmT}{\boldsymbol{\rm{T}}}

\newcommand{\bmf}{\boldsymbol{{ f }}}
\newcommand{\bmh}{\boldsymbol{{ h }}}

\newcommand{\bmEi}{\mathsf{E}}

\newcommand{\kdP}{\boldsymbol{{\mathcal{P}}}}
\newcommand{\kdF}{\boldsymbol{{\mathcal{F}}}}
\newcommand{\kdG}{\boldsymbol{{\mathcal{G}}}}
\newcommand{\kdH}{\boldsymbol{{\mathcal{H}}}}
\newcommand{\kdS}{\boldsymbol{{\mathcal{S}}}}
\newcommand{\kdK}{\boldsymbol{{\mathcal{K}}}}

\newcommand{\Diag}{\mathsf{Diag}}

\usepackage{amsmath,lipsum}
\usepackage{cuted}

\usepackage{amsmath,amssymb}    
\makeatletter
\renewcommand\normalsize{%
 \@setfontsize\normalsize\@xpt\@xiipt
 \abovedisplayskip 3\p@ \@plus1\p@ \@minus3\p@
 \abovedisplayshortskip \z@ \@plus1\p@
 \belowdisplayshortskip 1\p@ \@plus3\p@ \@minus1\p@
 \belowdisplayskip \abovedisplayskip
 \let\@listi\@listI}
\makeatother

\begin{document}
\author{Chungeng Tian\textsuperscript{\dag}, Ning Hao\textsuperscript{\dag}, Fenghua He\textsuperscript{*}, and Haodi Yao 
\thanks{\dag These authors contributed equally to this work.
The authors are with the National Key Laboratory of Modeling and Simulation
for Complex Systems, School of Astronautics, Harbin Institute of Technology,
Harbin 150001, China. (e-mail: tcghit@outlook.com, haoning0082022@163.com, hefenghua@hit.edu.cn, 20B904013@stu.hit.edu.cn).
}%
}
\maketitle
\thispagestyle{empty}
\pagestyle{empty}

\begin{abstract}
    This paper addresses the consistency issue of multi-robot distributed cooperative localization.  We introduce a consistent distributed cooperative localization algorithm conducting state estimation in a transformed coordinate.
    The core idea involves a linear time-varying coordinated transformation to render the propagation Jacobian independent of the state and make it suitable for a distributed manner.  This transformation is seamlessly integrated into a server-based distributed cooperative localization framework, in which each robot estimates its own state while the server maintains the cross-correlations. The transformation ensures the correct observability property of the entire framework. Moreover, the algorithm accommodates various types of robot-to-robot relative measurements, broadening its applicability. Through simulations and real-world dataset experiments, the proposed algorithm has demonstrated better performance in terms of both consistency and accuracy compared to existing algorithms.
\end{abstract}

\section{Introduction}

Cooperative localization (CL) is crucial for autonomous mobile robots during collaborative missions \cite{BorgeseTether-Based2022}, which collectively estimates the robot poses in a common reference frame
by utilizing proprioceptive sensors (e.g., wheel encoders or odometry) and exteroceptive sensors (e.g., cameras or laser scanners).
CL is resilient to environmental variations, such as scenarios with limited GPS signals or few landmarks. Nevertheless, in certain scenarios such as underwater or large-scale environments, communication may experience delays or be intermittently unavailable. Hence, the development of a distributed cooperative localization (DCL) framework that reduces communication costs is advantageous.

In DCL, two robot pose estimations are correlated after being updated with a relative measurement between them. If the pose estimations are continuously updated without considering these correlations, the problem of  {\it{double counting}} arises, leading to overconfidence in pose estimations \cite{bahrConsistentCooperativeLocalization2009}.

To address this issue, a widely used approach is the covariance intersection (CI) fusion technique \cite{julierNondivergentEstimationAlgorithm1997,carrillo-arceDecentralizedMultirobotCooperative2013,changResilientConsistentMultirobot2022,BiasedMeasurements}. However, CI is overly conservative because it assumes a maximal correlation between the estimations. To mitigate this conservatism, ellipsoidal intersection (EI) \cite{sijsStateFusionUnknown2012}, split covariance intersection (SCI) \cite{liSplitCovarianceIntersection2013}, and inverse covariance intersection (ICI) \cite{noackDecentralizedDataFusion2017,FDJLT} techniques have been proposed. Nonetheless, these methods are not suitable for handling partial observation cases, where the dimension of robot-to-robot relative measurements is smaller than the robot state. To address partial relative measurements, discorrelated minimum variance (DMV) \cite{zhuCooperativeLocalizationLimited2019}  integrate CI into the extended Kalman filter (EKF).
Unlike CI class methods, a more flexible optimization framework was proposed by constructing the unknown cross-covariance matrix instead of the upper bound of the whole joint covariance matrix.
The methods based on this framework, such as the game-theoretic approach \cite{leonardosGametheoreticApproachRobust2017} and the practical estimated minimum variance of cross-covariance (PECMV) \cite{chenCooperativeLocalizationUsing2022a}, exhibit less conservatism but require higher computational effort.
\begin{figure}[]
    \centering
    \includegraphics[width=0.48\textwidth]{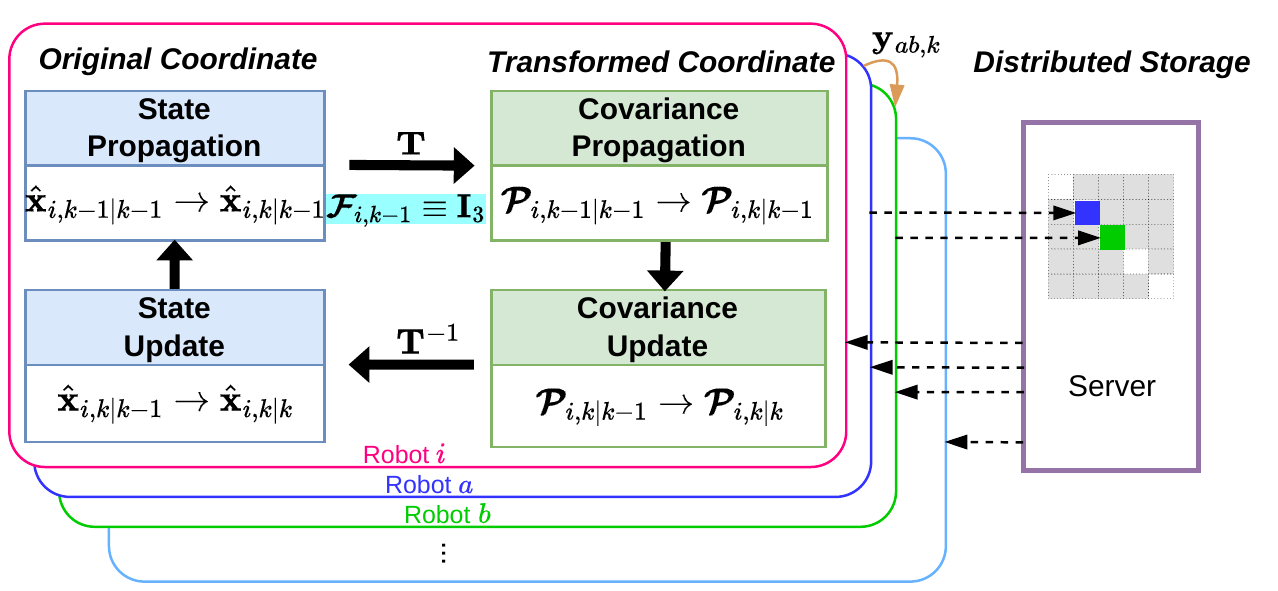}
    \caption{Transformed server-based DCL framework. The covariance matrices corresponding to the transformed error states propagate and update in the transformed coordinates.}
    \label{fig:cl_example}
    \vspace{-0.25cm}
\end{figure}

Another method used to tackle the double-counting issue is the central-equivalent EKF-based DCL. It considers all robot poses as the system state and uses a single filter to estimate the system state, automatically recording the correlations among robot pose estimations through cross-covariance matrices. Although central-equivalent DCL achieves high localization accuracy, its naive distributed form \cite{roumeliotisDistributedMultirobotLocalization2002} necessitates all-to-all communication for updating cross-covariance matrices. To alleviate communication costs, \cite{luftRecursiveDecentralizedLocalization2018} proposed a method that only requires pairwise communication between robots that obtain relative measurements. However, this method sacrifices estimation consistency since it updates the cross-covariance matrices with block-diagonal approximation matrices.
An alternative DCL algorithm was introduced in \cite{kiaCooperativeLocalizationMessage2015}. This algorithm incorporates a server responsible for managing and updating cross-covariance matrices. While this server-based framework enhances localization accuracy significantly, it also incurs substantial communication costs during the update steps. In a subsequent study \cite{kiaServerAssistedDistributedCooperative2018}, a technique equivalent to the Schmidt-Kalman filter is suggested to address issues related to unreliable communication links within the server-based framework.
While central-equivalent DCL and its distributed forms avoid double counting, they still suffer from inconsistency due to unjustified unobservable dimension reduction  \cite{huangAnalysisImprovementConsistency2008}  \cite{huangObservabilitybasedConsistentEKF2011}.
Since the absolute measurement is absent, the CL system is partially observable. \cite{KDIROS} has analytically proven that the EKF CL estimator exhibits an unobservable subspace of lower dimension than the original non-linear CL system. Consequently, the unjustified reduction of unobservable dimension leads to inconsistency and a decline in localization accuracy.

This paper focuses on enhancing the consistency of the central-equivalent DCL algorithms. We have devised a coordinate transformation that ensures the transformed propagation Jacobians are independent of the state. By incorporating this transformation into a server-based framework as shown in Figure \ref{fig:cl_example}, the correct observability property is guaranteed. With the assurance of correct observability, the estimator's consistency is enhanced. Furthermore, owing to the transformation, the transformed propagation Jacobian simplifies to an identity matrix, streamlining the propagation of cross-covariance matrices.

The main contributions of this paper are as follows.
\begin{itemize}
    \item A coordinate transformation is designed in light of decomposing the propagation Jacobian. The transformation renders the propagation Jacobian independent of the state.
    \item A consistent DCL algorithm is presented by integrating the transformation into a server-based framework, in which the state estimation is performed based on the transformed error states.
    \item The proposed DCL algorithm is validated by Monte Carlo simulations and experiments. The results show that it outperforms state-of-the-art algorithms in terms of consistency and accuracy.
\end{itemize}


\section {Problem description}
Consider a team of $N$ robots performing cooperative localization on a 2D plane in which each robot is equipped with proprioceptive sensors to sense its ego-motion information and exteroceptive sensors to obtain the robot-to-robot
relative measurements.

\subsection{Motion Model} Let $\bmx_{i,k} = \begin{bmatrix} \bmp_{i,k}^\top & \theta_{i,k}
    \end{bmatrix}^\top $  be the pose of robot $i$ in a common fixed frame at time step $k$, where $\bmp_{i,k} \in \mathbb{R}^2$ and $\theta_{i,k} \in \mathbb{R}$ are the position and
orientation of robot $i$, respectively, $i\in \{1,2,...,N\}$.
The motion of robot $i$ is described by
\begin{equation}
    \small
    \begin{array}{rl}
        \bmx_{i,k}
         & = \bmx_{i,k-1} +\begin{bmatrix}
            {\bmR}\left(\theta_{i,k-1}\right) & \bm{0}_{2\times 1} \\
            \bm{0}_{1\times 2}                & 1                  \\
        \end{bmatrix}({\bmu}_{i,k-1} \Delta t + \bm{\epsilon}_{i,k-1}) \\
         & \triangleq \bmf_i\left( \bmx_{i,k-1} ,{\bmu}_{i,k-1},\bm{\epsilon}_{i,k-1} \right)
    \end{array}
    \label{equ:motion1}%
\end{equation}
where ${\bmu}_{i,k-1} = \begin{bmatrix}
        \bmv_{i,k-1}^\top &
        \omega_{i,k-1}
    \end{bmatrix}^\top$ , $\bmv_{i,k-1} \in \mathbb{R}^2$ and $\omega_{i,k-1} \in \mathbb{R}$ are the linear and angular velocity of robot $i$,
$\bm \epsilon_{i,k-1} \in \mathbb{R}^{3}$ denotes white Gaussian noise with ${\mathbb{E}}(  \bm\epsilon_{i,k-1} \bm\epsilon_{i,k-1}^\top ) = \bmQ_{i,k-1}> \bm{0}$.
${\bmR} $ is the rotation matrix given by
$
    {\bmR}(\theta ) = \begin{bmatrix}
        \cos(\theta) & -\sin(\theta) \\
        \sin(\theta) & \cos(\theta)  \\
    \end{bmatrix}.
$

For the entire $N$ robot system, the state vector, the motion input vector, and the process noise vector are defined as the stacked vectors, i.e.,
$\small \bmx_k= \begin{bmatrix}
        \bmx_{1,k}^\top & \cdots & \bmx_{N,k}^\top
    \end{bmatrix}^\top$, $\small \bmu_{k-1} = \begin{bmatrix}\bmu_{1,k-1}^\top & \cdots & \bmu_{N,k-1}^\top
    \end{bmatrix}^\top$ and $\small \bm{\epsilon}_{k-1} =  \begin{bmatrix}
        \bm{\epsilon}_{1,k-1}^\top & \cdots & \bm{\epsilon}_{N,k-1}^\top
    \end{bmatrix}^\top$.
Then the system motion equation is given by
\begin{equation}
    \bmx_{k} = \bmf\left( \bmx_                      {k-1} ,{\bmu}_{k-1},\bm{\epsilon}_{k-1}\right).
    \label{equ:system_f}
\end{equation}

\subsection{Measurement Model}

At time step $k$, if robot $i,i\in\{1,2,...,N\}$ obtains a robot-to-robot relative measurement of robot $j,j\in\{1,2,...,N\}/i$, the relative measurement can be written as follows
\begin{equation}
    \bmy_{ij,k} =\bmh_{ij}(\bmx_{i,k},\bmx_{j,k})+\bm{\varepsilon}_{ij,k}                 \\
    \label{equ:measurement}
\end{equation}
where $\bm{\varepsilon}_{ij,k}$ is the measurement noise assumed to be the Gaussian white noise with $\bmEi(  \bm\varepsilon_{ij,k}  {\bm\varepsilon_{ij,k} }^\top ) = \bmR_{ij,k} > \bm{0}$.
To be consistent with our simulation and experiment, we assume that the measurements are relative position measurements
\begin{equation}
    \bmh_{ij}(\bmx_{i,k},\bmx_{j,k}) = {\bmR}(\theta_{i,k})^\top (\bmp_{j,k}-\bmp_{i,k}).
\end{equation}

Let $\{\dots,\bmy_{ij,k},\dots \}$ denote all the robot-to-robot relative measurements at time step $k$. Note that we do not require an all-to-all relative measurement topology in this paper.
The relative measurement vector of the system and the measurement noise vector are defined as the stacked vectors, i.e., $ \small \bmy_{k} = \begin{bmatrix}
        \cdots , {\bmy_{ij,k} }^\top , \cdots
    \end{bmatrix} ^\top$
and $\small \bm{\varepsilon}_{k} =  \begin{bmatrix}
        \cdots , {\bm{\varepsilon}_{ij,k} }^\top , \cdots
    \end{bmatrix} ^\top$.
Then, the system measurement equation is given by
\begin{equation}
    \bmy_{k} = \bmh(\bmx_k)+ \bm{\varepsilon}_{k}.
    \label{equ:system_h}
\end{equation}

\subsection{Linearized Error State Model}
Let $\hat{\bmx}_{k-1|k-1}$ denote the posterior estimate at time step $k-1$. Then the current time prior estimation $\hat{\bmx}_{k|k-1}$ is
\begin{equation}
    \hat{\bmx}_{k|k-1} = \bmf\left( \hat{\bmx}_{k-1|k-1} ,{\bmu}_{k-1},\bm{0} \right).
\end{equation}
Denoted the error state by $\tilde{\bmx}_{k} \triangleq \bmx_{k} - \hat{\bmx}_{k}$,  the linearized error state propagation equation for \eqref{equ:system_f} is
\begin{equation}
    \tilde{\bmx}_{k|k-1} =\bmF_{k-1}  \tilde{\bmx}_{k-1|k-1} + \bmG_{k-1} \bm{\epsilon}_{k-1}
    \label{sys:F}
\end{equation}
where $\bmF_{k-1}$ is the propagation Jacobian,
\begin{equation}
    \bmF_{k-1} =
    {\Diag}\left(
    {\bmF_{1,k-1} }, \dots , {\bmF_{N,k-1} }\right)
    \label{equ:F}
\end{equation}
\begin{equation}
    \bmG_{k-1}=
    {\Diag}\left(
    {\bmG_{1,k-1}}, \dots , {\bmG_{N,k-1} }\right)
\end{equation}
with
\begin{align}
    \bmF_{i,k-1 } & = \begin{bmatrix}
        \bmI_2           & \bmJ (\hat{\bmp}_{i,k|k-1} - \hat{\bmp}_{i,k-1|k-1}) \\
        \bm0_{1\times 2} & 1                                                    \\
    \end{bmatrix} \\
    \bmG_{i,k-1 } & = \begin{bmatrix}
        \bmR(\hat{\theta}_{k-1|k-1}) & \bm0_{2\times 1} \\
        \bm0_{1\times 2}             & 1                \\
    \end{bmatrix} %
\end{align}
and
$
    \bmJ = \begin{bmatrix}
        0 & -1 \\
        1 & 0
    \end{bmatrix}$.

Let $ \tilde{\bmy}_{k}\triangleq \bmy_{k} - \bmh (\hat{\bmx}_{k|k-1}  )$ denote the measurement residual vector, then the linearized measurement equation for \eqref{equ:system_h} is
\begin{equation}
    \tilde{\bmy}_{k} =  \bmH_k \tilde{\bmx}_{k|k-1} + \bm{\varepsilon}_{k},
    \label{sys:H}
\end{equation}
where $\bmH_k $ is the measurement Jacobian,
\begin{equation}
    \bmH_k =  \begin{bmatrix}
        \cdots & {\bmH_{ij,k} }^\top & \cdots
    \end{bmatrix}^\top
    \label{equ:H}
\end{equation}with
\begin{align}
    \bmH_{ij,k}       & = \begin{bmatrix}
        \cdots        &
        \bmH_{ij,k}^i & \cdots & \bmH_{ij,k}^j & \cdots
    \end{bmatrix}                                   \\
    {\bmH_{ij,k}^{i}} & = \bmR(\hat{\theta}_{i,k|k-1})^\top\begin{bmatrix}
        -\bmI_2 & -\bmJ(\hat{\bmp}_{j,k|k-1} - \hat{\bmp}_{i,k|k-1})
    \end{bmatrix}  \\
    {\bmH_{ij,k}^{j}} & = \bmR(\hat{\theta}_{i,k|k-1})^\top\begin{bmatrix}
        \bmI_2 & \bm0_{2\times 1}
    \end{bmatrix}.
\end{align}

\subsection{Observability}
\label{sec:analysis}
For estimators based on the linearized error state system \eqref{sys:F} and \eqref{sys:H},
the {\it local observability matrix} can be employed to perform the observability analysis\cite{149118}.  The local observability matrix over the time interval $[1,k]$ is given by
\begin{equation}
    \boldsymbol{\mathcal{O}} = \begin{bmatrix}
        {\bmH}_{1}            \\
        {\bmH}_{2} {\bmF}_{1} \\
        \vdots                \\
        {\bmH}_{k} {\bmF}_{k-1}  \cdots {\bmF}_{1}
    \end{bmatrix}.
    \label{equ:O}
\end{equation}
An observability matrix for $N$ robots takes up too much space and is hard to read. Following \cite{huangAnalysisImprovementConsistency2008}, we
consider a simple case in which the robot number is limited to 2 and only robot 1 can observe robot 2.
Note that such a simple scenario is sufficient to illustrate the inconsistency problem.
The local observability matrix for this case is written in \eqref{equ:sys_O} where
\begin{figure*}
    \centering
    \begin{equation}
        \boldsymbol{\mathcal{O}} = \bmM \begin{bmatrix}
            -\bmI_2 & -\bmJ (\hat{\bmp}_{2,1|0} - \hat{\bmp}_{1,1|0})                         & \bmI_2 & \bm0_{2\time 1}                                                        \\
            -\bmI_2 & -\bmJ (\hat{\bmp}_{2,2|1} - \hat{\bmp}_{1,1|0}) + \delta \bmp_{1,1}     & \bmI_2 & \bmJ (\hat{\bmp}_{2,2|1} - \hat{\bmp}_{2,1|0}) -\delta \bmp_{2,1}      \\
            \vdots  & \vdots                                                                  & \vdots & \vdots                                                                 \\
            -\bmI_2 & -\bmJ (\hat{\bmp}_{2,k|k-1} - \hat{\bmp}_{1,1|0}) + \delta \bmp_{1,k-1} & \bmI_2 & \bmJ (\hat{\bmp}_{2,k|k-1} - \hat{\bmp}_{2,1|0}) - \delta \bmp_{2,k-1} \\
        \end{bmatrix}
        \label{equ:sys_O}
    \end{equation}
    \hrulefill
\end{figure*}
\begin{equation}
    \bmM = \text{Diag} \big(
    \bmR(\hat\theta_{1,1|0}), \bmR(\hat\theta_{1,2|1}), \dots,  \bmR(\hat\theta_{1,k|k-1})
    \big)
\end{equation}
\begin{equation}
    \delta \bmp_{i,k} = \sum_{l=1}^{k} \hat{\bmp}_{i,l|l} -\hat{\bmp}_{i,l|l-1}.
    \label{equ:res}
\end{equation}

One can see that if $\delta \bmp_{i,k}$ is zero, there exist three unobservable dimensions in CL, s.t. $ \boldsymbol{\mathcal{O}}\bmN= \bm0$ with unobservable subspace
\begin{equation}
    \bmN = \begin{bmatrix}
        \bmI_2           & \bmJ \hat{\bmp}_{1,1|0} \\
        \bm0_{1\times 2} & 1                       \\
        \bmI_2           & \bmJ \hat{\bmp}_{2,1|0} \\
        \bm0_{1\times 2} & 1                       \\
    \end{bmatrix}.
\end{equation}
The presence of three unobservable dimensions in the CL system arises from the inability to estimate the global position (x, y) and orientation (yaw) without absolute measurements.
However, $\delta \bmp_{i,k} = \bf0$ is almost impossible to hold, resulting in an erroneous reduction of the unobservable subspace dimension to 2, as follows:
\begin{equation}
    \bmN = \begin{bmatrix}
        \bmI_2
         & \bm0_{2\times 1}
         & \bmI_2
         & \bm0_{2\times 1}
    \end{bmatrix}^\top.
\end{equation}
Correspondingly, the global orientation is erroneously observable.

Since central-equivalent DCL algorithms are also based on \eqref{sys:F} and \eqref{sys:H},
they will be inconsistent due to the reduction of the unobservable dimension, which in turn leads to a decrease in accuracy.

\section{Coordinate Transformation Based DCL}
In this section, we design a coordinate transformation to ensure the correct observability property of the central-equivalent DCL algorithms.
\subsection[]{Transformed  Linearized Error State System}
Section \ref{sec:analysis} reveals that the propagation Jacobian, which depends on the state estimates, introduces inconsistencies in central-equivalent DCL. More specifically, the difference between prior and posterior estimates results in $\delta \mathbf{p}_{i,k} \neq \mathbf{0}$, which subsequently leads to an unjustified reduction in unobservable dimensions.

An approach to address this inconsistency issue is to render the propagation Jacobian independent of the state.
The state-independent propagation Jacobian can prevent the inconsistent terms like \eqref{equ:res} appear in the local observability matrix, thus automatically guaranteeing the correct observability.
Fortunately, employing state estimation in a transformed coordinate system \cite{KDIROS} facilitates the achievement of this objective, in which the state variables in
propagation Jacobian can be transmitted to the transformation matrices.
Let the coordinate transformation $\bmT_k$ be a time-varying matrix, and $\bmz_k$ be the error state in the transformed coordinate, which is obtained by multiplying $\bmT_k$ with the error state $\tilde{\bmx}_k$:
\begin{equation}
    \bmz_{k|k-1} = \bmT_{k|k-1}\tilde{\bmx}_{k|k-1},\quad  \bmz_{k|k} = \bmT_{k|k}\tilde{\bmx}_{k|k},
    \label{equ:z}
\end{equation}
where $\bmT_{k|k-1}$ and $ \bmT_{k|k}$ are evaluated at $\hat{\bmx}_{k|k}$ and $\hat{\bmx}_{k|k-1}$, respectively.
Substituting
\eqref{equ:z} into \eqref{sys:F} and \eqref{sys:H} yields the linearized error state system in the transformed coordinate
\begin{equation}
    \setlength{\arraycolsep}{1.5pt}
    \left\{
    \begin{array}{rcl}
        {\bmz}_{k|k-1}   & = & \kdF_{k-1}  {\bmz}_{k-1|k-1}  + \kdG_{k-1} \bm{\epsilon}_{k-1} \\
        \tilde{\bmy}_{k} & = & \kdH_k {\bmz}_{k|k-1} + \bm{\varepsilon}_{k}                   %
    \end{array}
    \right.
    \label{sys:FHz}
\end{equation}
where
\vspace{-0.2cm}
\begin{align}
    \kdF_{k-1} & = \bmT_{k|k-1} \bmF_{k-1} {\bmT_{k-1|k-1}^{-1}} ,\label{equ:TFT} \\
    \kdG_{k-1} & = \bmT_{k|k-1}\bmG_{k-1}, \label{equ:TG}                         \\
    \kdH_k     & =  \bmH_k  {\bmT_{k|k-1}^{-1}} .\label{equ:HT}
\end{align}

Not all transformations can render the transformed propagation Jacobian independent of the state.
For example, the trivial transformation $\mathbf{T}_k = \mathbf{I}_{3N}$ does not take any effect.
There are two requirements for designing the transformation:
\begin{itemize}
    \item The transformation can render the transformed propagation Jacobian independent of the state to ensure the correct observability.
    \item The transformation matrix is a diagonal block matrix. Otherwise, the state estimates of different robots would be coupled to each other, which is unfavorable for DCL.
\end{itemize}
Consequently, it is a challenging work to design such a suitable transformation.
Our design is inspired by the decomposition of the original propagation Jacobian as
\begin{align}
    \setlength{\arraycolsep}{0.03cm}
    \bmF_{i,k-1} & = \begin{bmatrix}
        \bmI_2           & \bmJ \hat{\bmp}_{i,k|k-1} \\
        \bm0_{1\times 2} & 1                         \\
    \end{bmatrix}
    \setlength{\arraycolsep}{0.05cm}
    \begin{bmatrix}
        \bmI_2           & - \bmJ \hat{\bmp}_{i,k-1|k-1} \\
        \bm0_{1\times 2} & 1                             \\
    \end{bmatrix}.
    \label{equ:Fi_dec}
\end{align}
By comparing \eqref{equ:Fi_dec} with \eqref{equ:TFT}, we can construct a proper transformation matrix:
\begin{equation}
    \bmT_k = \Diag(\bmT_{1,k}, ...,\bmT_{N,k} )
    \label{equ:Tk}
\end{equation}
with
\begin{equation}
    \bmT_{i,k}=  \begin{bmatrix}
        \bmI_2           & -\bmJ \hat{\bmp}_{i,k} \\
        \bm0_{1\times 2} & 1                      \\
    \end{bmatrix},   \bmT_{i,k}^{-1}=  \begin{bmatrix}
        \bmI_2           & \bmJ \hat{\bmp}_{i,k} \\
        \bm0_{1\times 2} & 1                     \\
    \end{bmatrix}.
\end{equation}
Then, we have
\begin{align}
    \kdF_{k-1} & = \bmI_{3N}                                 \\
    \kdG_{k-1} & = \Diag({\kdG_{1,k-1},\dots ,\kdG_{N,k-1}}) \\
    \kdH_{k}   & =   \begin{bmatrix}
        \cdots & {\kdH_{ij,k} }^\top & \cdots
    \end{bmatrix}^\top
\end{align}
with
\begin{align}
    \kdG_{i,k-1}  & =  \begin{bmatrix}
        \bmR(\hat{\theta}_{i,k-1|k-1}) & -\bmJ \hat{\bmp}_{i,k|k-1} \\
        \bm0_{1\times 2}               & 1                          \\
    \end{bmatrix}                           \label{equ:Gi} \\
    \kdH_{ij,k}   & =  \begin{bmatrix}
        \cdots        &
        \kdH_{ij,k}^i & \cdots & \kdH_{ij,k}^j & \cdots
    \end{bmatrix}                                          \\
    \kdH_{ij,k}^i & =  \bmR(\hat{\theta}_{i,k|k-1})^\top\begin{bmatrix}
        - \bmI_2 & - \bmJ\hat{\bmp}_{j,k|k-1}
    \end{bmatrix}         \\
    \kdH_{ij,k}^j & =  \bmR(\hat{\theta}_{i,k|k-1})^\top\begin{bmatrix}
        \bmI_2 & \bmJ\hat{\bmp}_{j,k|k-1}
    \end{bmatrix}.
\end{align}
Now we get a state-independent transformed propagation Jacobian $\kdF_{k-1}$.
The more interesting property is that it's also an identity matrix, which is more suitable for DCL than the original propagation Jacobian as shown in Section \ref{sec:TDCL}.

\subsection{ Transformed Server-based DCL}
\label{sec:TDCL}
We deploy the transformation on a server-based framework proposed in \cite{kiaCooperativeLocalizationMessage2015,kiaServerAssistedDistributedCooperative2018}.
The server-based DCL is a central-equivalent method, in which each robot propagates its estimations while the server maintains the cross-correlations.
Note that besides the server-based framework, the transformation can be also integrated into other central-equivalent DCL algorithms. We choose the server-based framework for its better performance.
The steps of the proposed transformed server-based DCL are shown in Algorithm \ref{alg:d_Transformed_EKF}.
\begin{algorithm}
    \label{alg:d_Transformed_EKF}
    \caption{Transformed Server-based DCL}
    \textbf{Propagation:}\\
    \quad \textbf{\color{gray!90} Server:} \\
    \quad \quad{\small  $\kdP_{ij,k|k-1} \gets \kdP_{ij,k-1|k-1}  \quad i\neq j \in \{1,\dots,N\}$} \\
    \quad \textbf{\color{gray!90} Robot i:}\\
    \quad \quad {\small propagate $\hat{\bmx}_{i,k|k-1} $ using  \eqref{equ:prof} }\\
    \quad \quad {\small propagate $\kdP_{i,k|k-1} $ using  \eqref{equ:proP} }\\

    \textbf{Update:}\\
    \quad \tcp{measurement $\bmy_{ab,k}$}
    \quad \textbf{\color{gray!90} Server:}\\
    \quad \quad {\small receive $ (\hat{\bmx}_{a,k|k-1},\kdP_{a,k|k-1}$,$\bmy_{a b,k})$ from robot $a$} \\
    \quad \quad {\small receive $ (\hat{\bmx}_{b,k|k-1},\kdP_{b,k|k-1})$ from robot $b$}\\
    \quad \quad {\small calculate  $(\bm{r}_i, \boldsymbol{\varGamma }_i)$ for each robot using \eqref{equ:c1} and \eqref{equ:c2} }\\
    \quad \quad {\small update $\kdP_{ij,k|k}$ using \eqref{equ:kdpij}} \\
    \quad \textbf{\color{gray!90} Robot i:}\\
    \quad \quad {\small receive $(\bm{r}_i, \boldsymbol{\varGamma }_i)$ from Server} \\
    \quad \quad {\small update $\hat{\bmx}_{i,k|k}$ using \eqref{equ:bmz_update} } \\
    \quad \quad {\small update ${\kdP}_{i,k|k}$ using  \eqref{equ:bmz_update2}} \\
\end{algorithm}

\subsubsection{Propagation in the transformed system}
Different from the original server-based DCL, the proposed algorithm estimates the robot states based on the transformed error state.
Robot $i$ maintains $\hat{\bmx}_{i,k}$ and its transformed covariance $\kdP_{i,k}$, while the server retains the transformed cross-covariances $\{\kdP_{ij}\}_{i,j\in\{1,\dots,N\}}^{i\neq j}$,
where
\begin{align}
    \kdP_{i,k}  & \triangleq \bmT_{i,k}  \bmP_{i,k}\bmT_{i,k}^\top     \\
    \kdP_{ij,k} & \triangleq \bmT_{i,k}  \bmP_{ij,k}\bmT_{j,k}^\top  .
\end{align}
The state estimates and the transformed covariance matrix propagates as
    {\small
        \begin{align}
            \hat{\bmx}_{i,k|k-1} & = \bmf_i(\hat{\bmx}_{i,k-1|k-1},\bmu_{k-1},\bf0 )           \label{equ:prof}                          \\
            \kdP_{i,k|k-1}       & ={\kdF}_{i,k-1} \kdP_{i,k-1|k-1}{\kdF}_{i,k-1}^\top + {\kdG}_{i,k-1} \bmQ_{i,k-1} {\kdG}_{i,k-1}^\top
            \label{equ:proP}.
        \end{align}
    }
As the transformed propagation Jacobian $\kdF_{i,k-1}$ is an identity matrix, the transformed cross-covariances stored in the server do not need to change in the propagation steps.
\begin{remark}
    Compared to the original server-based DCL, the cross-covariances do not need to propagate. This nature allows us not to design additional mechanisms for propagation. The transformation can be also utilized in other DCL algorithms to reduce the complexity of the cross-covariance propagation.
\end{remark}

\subsubsection{Update in the transformed system}

Assume that robot $ a$ obtains a  relative measurement $\bmy_{a b,k}$
on robot $b$ at time step $k$.  The update steps are detailed below.

Firstly, the server receives a message $( \hat{\bmx}_{a,k|k-1},\kdP_{a,k|k-1}$,$\bmy_{a b,k})$ from robot $a$ and a message $ (\hat{\bmx}_{b,k|k-1},\kdP_{b,k|k-1})$ from robot $b$.

Subsequently, the server sends the correction message to each robot. The correction message $(\bm{r}_i, \boldsymbol{\varGamma }_i)$ for robot $i$ is
\begin{align}
    \bm{r}_i                  & = \kdK_i (\bmy_{ab,k} - {\bmh}_{a b}(\hat{\bmx}_{a,k|k-1},\hat{\bmx}_{b,k|k-1})) \label{equ:c1} \\
    \boldsymbol{\varGamma }_i & = \kdK_i \kdS \kdK_i^\top \label{equ:c2}
\end{align}
where \begin{equation}
    \kdK_i =  (\kdP_{i a,k|k-1} {{\kdH}_{ab,k}^{a}}^\top +  \kdP_{i b,k|k-1} {{\kdH}_{ab,k}^{b}}^\top  )\kdS^{-1}\\
    \label{equ:Ki}
\end{equation} is the transformed Kalman gain obtained by minimizing the trace of the transformed covariance,
\begin{equation}
    \setlength{\arraycolsep}{1pt}
    \begin{array}{rl}
        \kdS & =  \kdH_{ab,k}^{a} \kdP_{a,k|k-1} {\kdH_{ab,k}^{a}}^\top +  \kdH_{ab,k}^{a} \kdP_{a b,k|k-1} {\kdH_{ab,k}^{b}}^\top       \\
             & \quad +  \kdH_{ab,k}^{b} \kdP_{b a,k|k-1}  {\kdH_{ab,k}^{a}}^\top + \kdH_{ab,k}^{b} \kdP_{b,k|k-1} {\kdH_{ab,k}^{b}}^\top \\
             & \quad + \bmR_{a b ,k}.
    \end{array}
    \label{equ:Sab}
\end{equation}

Lastly, each robot updates its estimates with the correction message.
\begin{align}
    \hat{\bmx}_{i,k|k} & = \hat{\bmx}_{i,k|k-1} + \bmT_{i,k|k-1}^{-1}  \bm{r}_i \label{equ:bmz_update} \\
    \kdP_{i,k|k}       & =  \kdP_{i,k|k-1} - \boldsymbol{\varGamma }_i.  \label{equ:bmz_update2}
\end{align}
As the transformed Kalman gain minimizes the trace of the transformed covariance, the estimate correction $\bm{r}_i$ is also expressed in the transformed coordinate system.
The reverse transformation of $\bm{r}_i$ into the original coordinate system is required, as indicated in \eqref{equ:bmz_update}.
Simultaneously, the transformed cross-covariances stored in the server are updated as:
\begin{equation}
    \kdP_{ij,k|k} = \kdP_{ij,k|k-1} -   \kdK_i \kdS \kdK_j^\top,  {i\neq j\in\{1,\dots,N\}}.
    \label{equ:kdpij}
\end{equation}

\begin{remark}
    As the robot number increases, the communication cost of updates can be significant. In \cite{kiaServerAssistedDistributedCooperative2018}, a method equivalent to the Schmidt-Kalman filter \cite{schmidt1966application} is developed to deal with the message dropouts. Furthermore, we can use this method to actively reduce the communication cost. For example, instead of all robots, only robot $a$ and robot $b$ update their estimates given relative measurement $\bmy_{ab,k}$. Correspondingly, the server only updates the cross-covariances related to robot $a$ and robot $b$.  For more information, the reader is recommended to refer to \cite{kiaServerAssistedDistributedCooperative2018}.
\end{remark}

\subsection{Observability Property}
The local observability matrix for the transformed system \eqref{sys:FHz} over the time interval $[1,k]$ is given by
\begin{equation}
    \boldsymbol{\mathcal{O}}' = \begin{bmatrix}
        {\kdH}_{1}            \\
        {\kdH}_{2} {\kdF}_{1} \\
        \vdots                \\
        {\kdH}_{k} {\kdF}_{k-1}  \cdots {\kdF}_{1}
    \end{bmatrix}.
    \label{equ:O_trans}
\end{equation}
As the same as \ref{sec:analysis}, we consider a simple case with only two robots. The observability matrix for the simple case is written as
\begin{equation}
    \boldsymbol{\mathcal{O}}' = \bmM \begin{bmatrix}
        -\bmI_2 & -\bmJ \hat \bmp_{2,1|0}  & \bmI_2 & \bmJ \hat\bmp_{2,1|0}   \\
        -\bmI_2 & -\bmJ \hat \bmp_{2,2|1}  & \bmI_2 & \bmJ \hat\bmp_{2,2|1}   \\
        \vdots  & \vdots                   & \vdots & \vdots                  \\
        -\bmI_2 & -\bmJ\hat \bmp_{2,k|k-1} & \bmI_2 & \bmJ \hat\bmp_{2,k|k-1}
    \end{bmatrix}.
\end{equation}
There exists three  unobservable dimensions for the transformed system, s.t. $\boldsymbol{\mathcal{O}}' \bmN' = \bm0$, with unobservable subspace
\begin{equation}
    \bmN' = \begin{bmatrix}
        \bmI_3 & \bmI_3
    \end{bmatrix}^\top.
\end{equation}
Thus, the proposed transformed server-based DCL has the correct observability property.

\section[]{Monte Carlo Simulations}
A series of Monte Carlo simulations were conducted to compare the proposed algorithm with other algorithms under various conditions, aiming to validate the preceding observability property analysis and demonstrate the capabilities of the proposed algorithm.
\subsection{Setup}
\begin{figure}[htbp]
    \centering
    \includegraphics[width = 0.45 \textwidth]{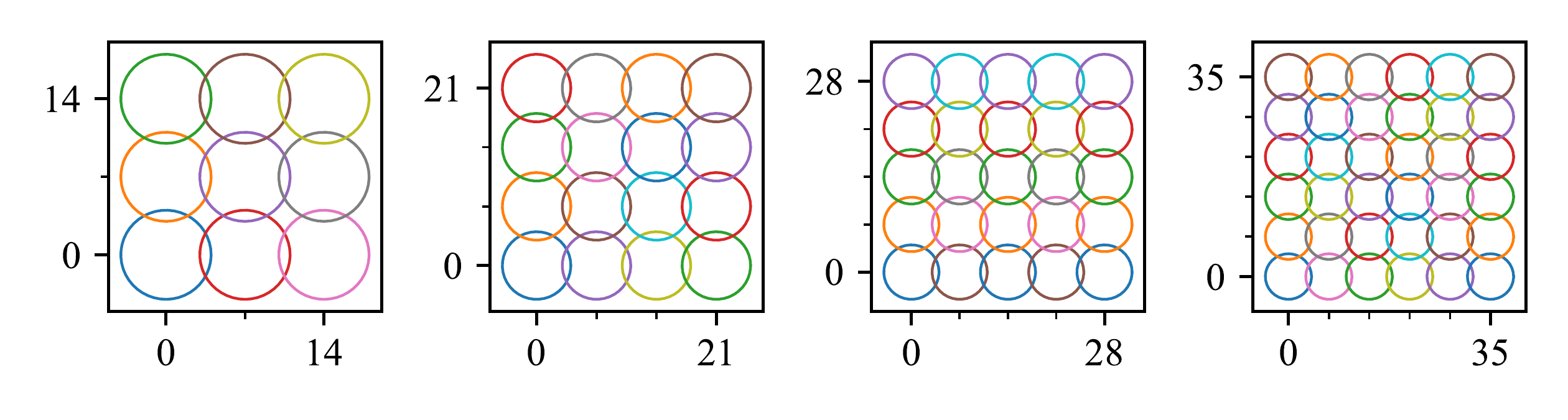}
    \vspace{-0.3cm}
    \caption{Simulated trajectories of 9, 16, 25, and 36 robots. The unit is meter. The robots move in a circle with a radius of four meters. The period of one circle of motion varies from 20 to 40 seconds. }
    \label{fig:traj}
\end{figure}

In the simulation scenario, robots moved 360 seconds on a square field.
The robot trajectories are shown in Figure \ref{fig:traj}.
The linear velocities $\bmv_i$ of the robots vary between 0.628 m/s and 1.256 m/s, with angular velocities $\omega_i$ ranging from 0.157 rad/s to 0.314 rad/s. The odometry noise $\bm \epsilon _{i}$ is [0.02 m, 0 m, 0.005 rad], and the time interval for odometry calculations $\Delta t$ is fixed at 0.1 seconds. Inter-robot relative position information is obtained through a combination of relative bearing and distance measurements. The noise levels associated with bearing and distance measurements are 0.01 rad and 0.2 m, respectively, with relative position measurements being conducted at a frequency of 2 Hz.

We compared the following methods in the Monte Carlo simulations:
\begin{itemize}
    \item Central:
          A centralized Cooperative Learning (CL) algorithm that utilizes a joint Extended Kalman Filter (EKF)\cite{roumeliotisDistributedMultirobotLocalization2002}. However, it calculates the Jacobians using ground truth instead of state estimates to ensure correct observability \cite{huangObservabilitybasedConsistentEKF2011}. This method serves as the baseline.
    \item OSB: The original server-based DCL algorithm \cite{kiaServerAssistedDistributedCooperative2018}.
    \item TSB: The proposed transformed server-based DCL algorithm.
\end{itemize}
Although this paper does not specifically investigate the impact of communication failures on the performance of DCL algorithms, for the reader's reference, we assess the performance of the DCL algorithms under three communication conditions: a 50\% success rate (OSB-50 and TSB-50), a 75\% success rate (OSB-75 and TSB-75) and a 99\% success rate (OSB-99 and TSB-99).
Each algorithm undergoes 100 Monte Carlo runs.

\subsection{Metrics}
The evaluation metrics used in this paper comprise the root mean square error (RMSE) for accuracy assessment and the normalized estimation error squared (NEES) for consistency evaluation. A lower RMSE indicates higher precision, while a NEES value closer to 1 signifies better consistency.

\subsection{Results}
\subsubsection{Test on a fixed number of robots and various sensor ranges}

\begin{figure}[htbp]
    \centering
    \includegraphics[width = 0.435 \textwidth]{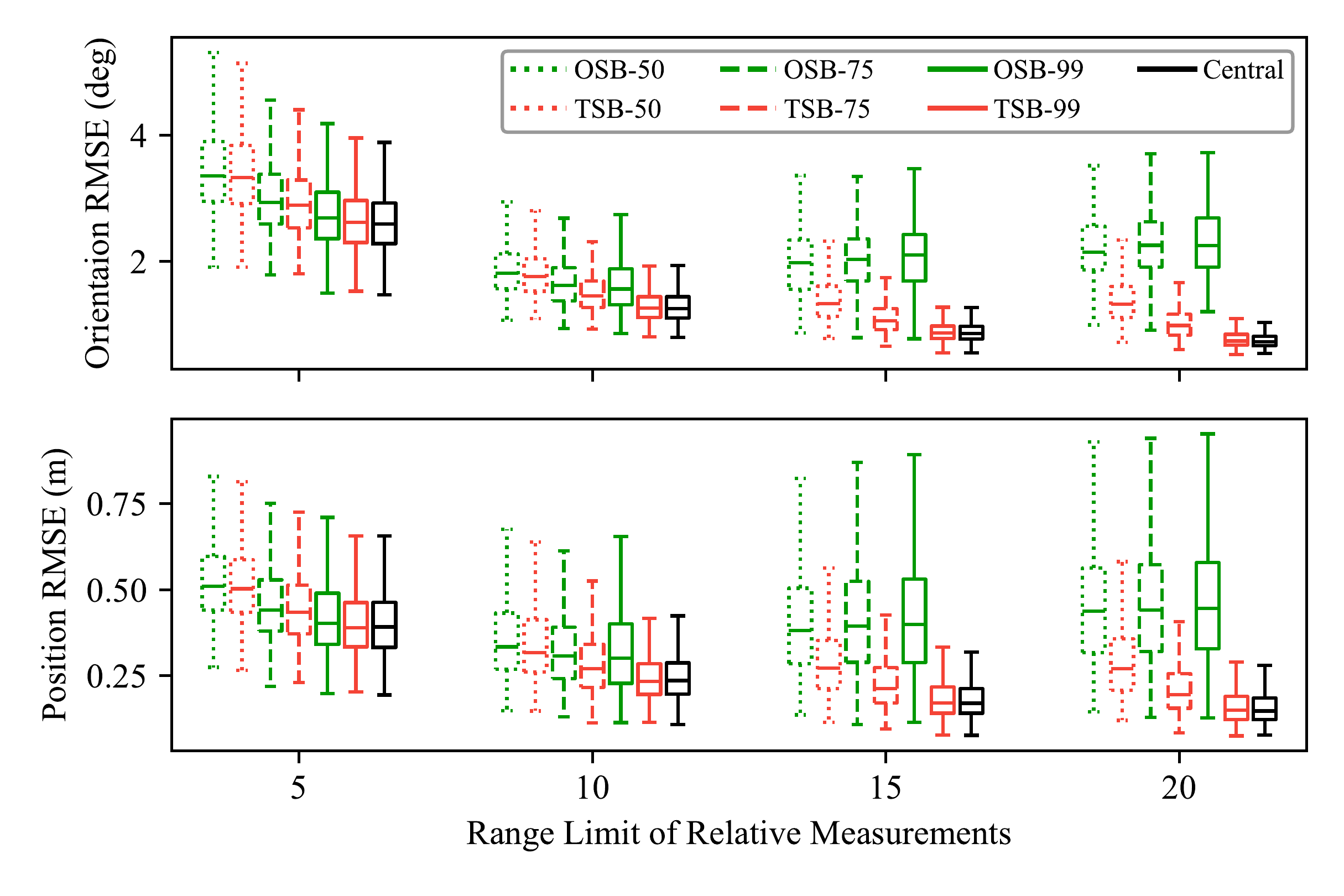}
    \vspace{-0.3cm}
    \caption{ Orientation and position RMSE with $N=16$ in 100 Monte Carlo runs with different sensor range limits.}
    \label{fig:dis_thr}
\end{figure}

\begin{figure}[htbp]
    \centering
    \includegraphics[width = 0.435 \textwidth]{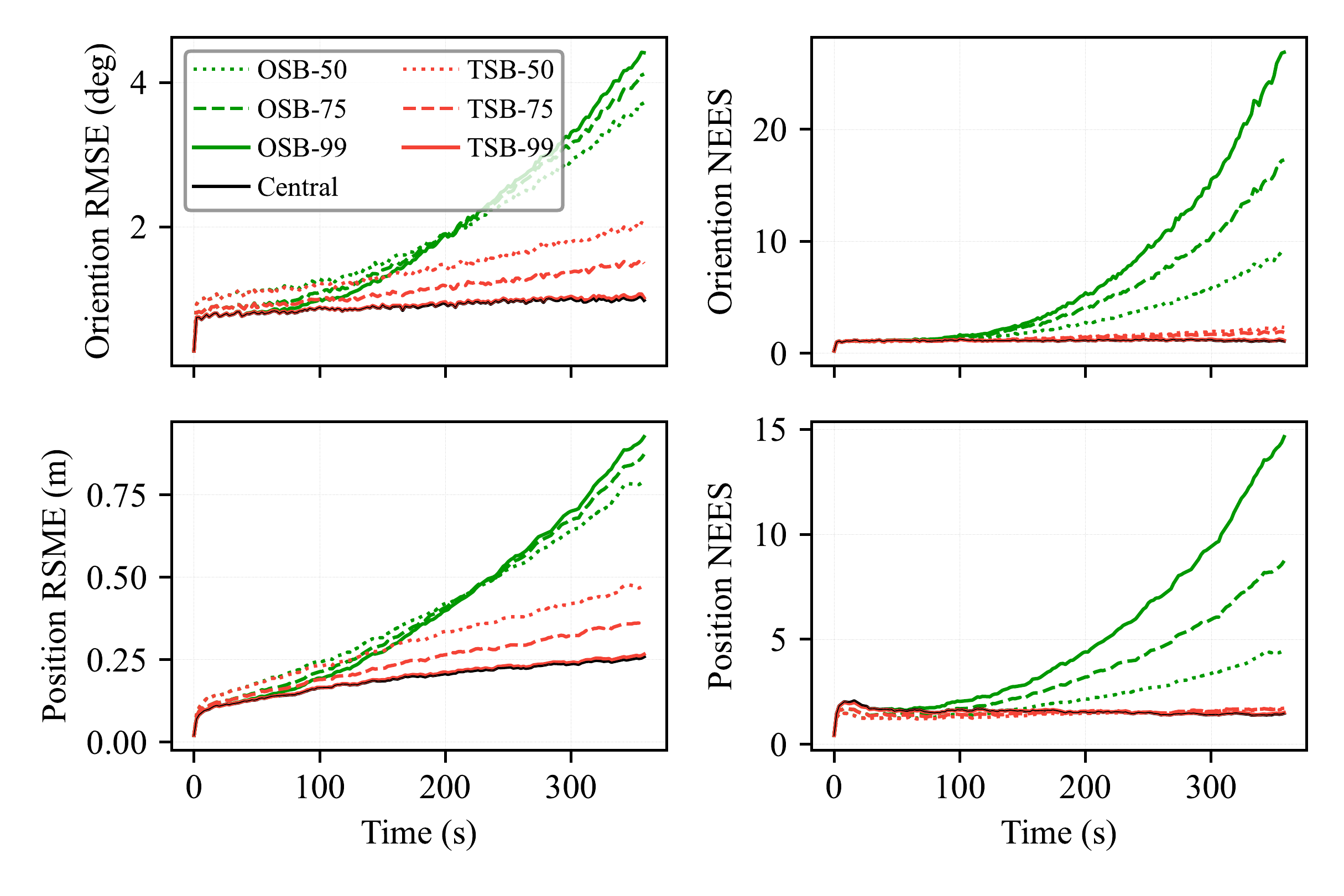}
    \vspace{-0.35cm}
    \caption{Average RMSE and NEES of 100 Monte Carlo simulations over time with $N=16$ and sensor range limited to 15 m.
    }
    \vspace{-0.05cm}
    \label{fig:n4}
\end{figure}

In the first test, the robot count was fixed at $N=16$, and the sensor range was limited to 5, 10, 15, or 20 meters. Figure \ref{fig:dis_thr} reports the position and orientation  RMSE  of 10 robots across 100 Monte Carlo simulations.
Table \ref{table:dis_thr} presents the average NEES of these algorithms.
\begin{table}[ht]
    \centering
    \tabcolsep=0.18cm
    \caption{Average  Orientation / Position  NEES of 100 Monte Carlo Simulations with $N=16$ }
    \label{table:dis_thr}
    \begin{tabular}{ccccc}
        \toprule
        Sensor range & 5 m                           & 10 m                          & 15 m                          & 20 m                          \\
        \midrule[0.5pt]
        Central      & 1.02 / 1.02                   & 1.08 / 1.47                   & 1.13 / 1.56                   & 1.09 / 1.53                   \\
        \midrule[0.1pt]
        OSB-50       & 1.07 / 1.06                   & 1.30 / 1.31                   & 2.99 / 2.18                   & 5.01 / 3.05                   \\
        TSB-50       & \textbf{1.02} / \textbf{1.01} & \textbf{1.15} / \textbf{1.21} & \textbf{1.44} / \textbf{1.41} & \textbf{1.89} / \textbf{1.63} \\
        \midrule[0.1pt]
        OSB-75       & 1.13 / 1.08                   & 1.56 / 1.61                   & 4.81 / 3.37                   & 7.84 / 4.83                   \\
        TSB-75       & \textbf{1.03} / \textbf{1.00} & \textbf{1.12} / \textbf{1.33} & \textbf{1.34} / \textbf{1.50} & \textbf{1.55} / \textbf{1.57} \\
        \midrule[0.1pt]
        OSB-99       & 1.18 / 1.12                   & 1.95 / 2.02                   & 6.68 / 4.90                   & 10.7 / 7.21                   \\
        TSB-99       & \textbf{1.04} / \textbf{1.00} & \textbf{1.08} / \textbf{1.42} & \textbf{1.14} / \textbf{1.56} & \textbf{1.14} / \textbf{1.52} \\
        \bottomrule
    \end{tabular}
\end{table}
As the range limit increases,
both Central and the proposed method OSB demonstrate better accuracy.
However, when the range limit exceeds 10 meters, the performance of the OSB methods deteriorates.
We attribute the inferior performance of OSB to overconfidence in orientation estimation as discussed in Section \ref{sec:analysis}.
As the range limit and communication success rate rise, more relative measurements become accessible for updating state estimates. This leads to a significant decrease in the consistency of OSB, as demonstrated in Table \ref{table:dis_thr}.
Over time, the consistency of OSB declines further compared to TSB, as depicted in Figure \ref{fig:n4}.

\subsubsection{Test on a fixed sensor range and various number of robots}
\begin{figure}[hbp]
    \centering
    \includegraphics[width = 0.435 \textwidth]{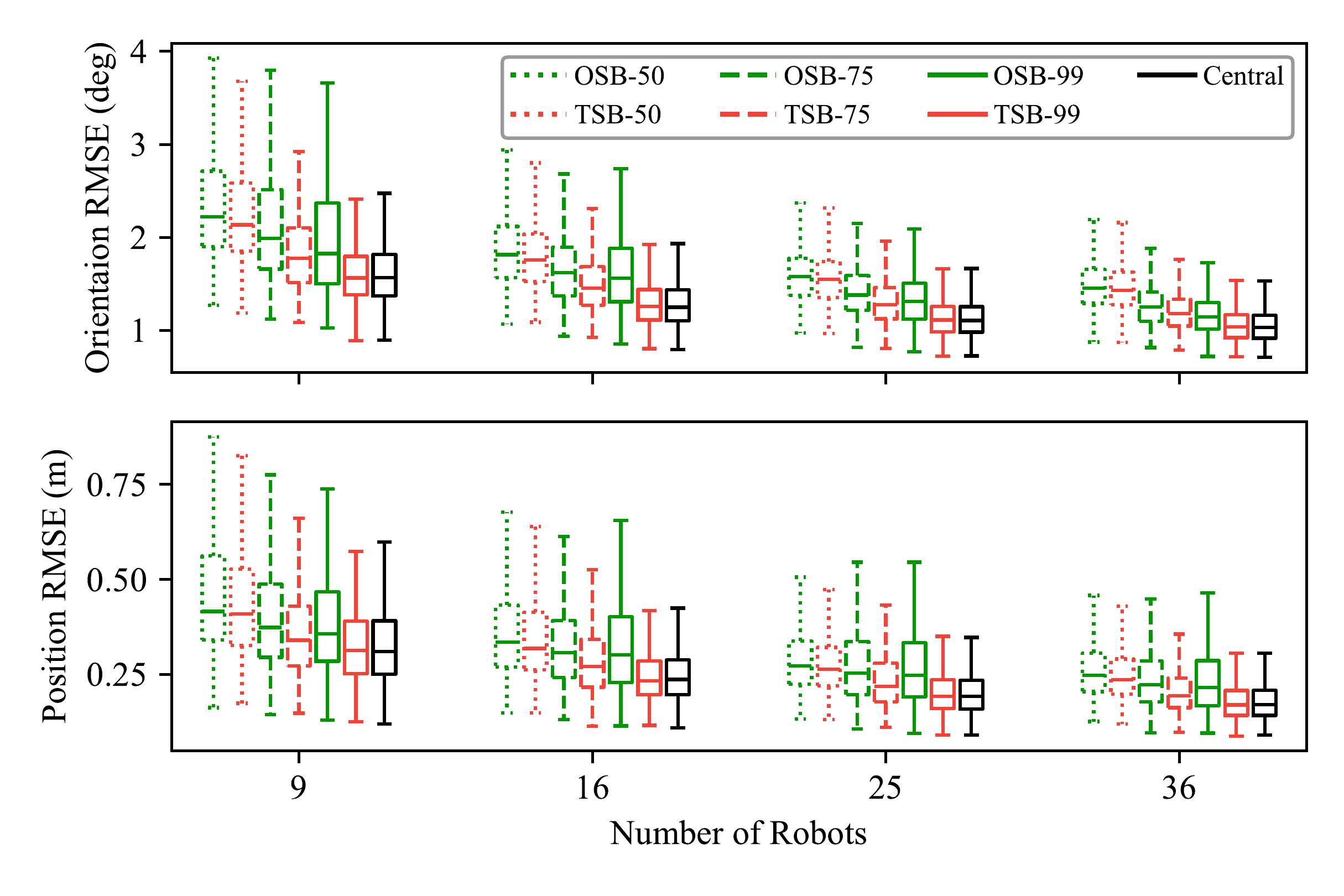}
    \vspace{-0.35cm}
    \caption{ Orientation and position RMSE of $N$ robots in 100 Monte Carlo runs with  sensor range limited to 10 m.}
    \vspace{-0.05cm}
    \label{fig:robot_num}
\end{figure}

In the second test, the sensor range is restricted to 10 meters. The algorithms are evaluated with various numbers of robots. Figure \ref{fig:robot_num} illustrates the average RMSE of all robot estimates in 100 Monte Carlo simulations as the number of robots is raised from 9 to 36.
Table \ref{table:robot_num} reports the average NEES of these algorithms.
As the number of robots increases, both OSB and TSB demonstrate enhanced accuracy. However, the proposed algorithm TSB still outperforms OSB in terms of accuracy and consistency.
\begin{table}[htbp]
    \centering
    \caption{Average  Orientation / Position  NEES of 100 Monte Carlo Simulations with sensor range limited to 10 m }
    \label{table:robot_num}
    \tabcolsep=0.18cm
    \begin{tabular}{ccccc}
        \toprule
        Robot number & 9                             & 16                            & 25                            & 36                            \\
        \midrule[0.5pt]
        Central      & 1.07 / 1.31                   & 1.08 / 1.47                   & 1.08 / 1.39                   & 1.08 / 1.46                   \\
        \midrule[0.1pt]
        OSB-50       & 1.38 / 1.26                   & 1.30 / 1.31                   & 1.12 / 1.18                   & 1.10 / 1.27                   \\
        TSB-50       & \textbf{1.11} / \textbf{1.16} & \textbf{1.15} / \textbf{1.21} & \textbf{1.05} / \textbf{1.11} & \textbf{1.05} / \textbf{1.19} \\
        \midrule[0.1pt]
        OSB-75       & 1.62 / 1.37                   & 1.56 / 1.61                   & 1.34 / 1.59                   & 1.21 / 1.56                   \\
        TSB-75       & \textbf{1.05} / \textbf{1.15} & \textbf{1.12} / \textbf{1.33} & \textbf{1.08} / \textbf{1.29} & \textbf{1.06} / \textbf{1.26} \\
        \midrule[0.1pt]
        OSB-99       & 1.87 / 1.63                   & 1.95 / 2.02                   & 1.63 / 2.06                   & 1.37 / 2.10                   \\
        TSB-99       & \textbf{1.04} / \textbf{1.30} & \textbf{1.08} / \textbf{1.42} & \textbf{1.07} / \textbf{1.37} & \textbf{1.07} / \textbf{1.43} \\
        \bottomrule
    \end{tabular}
\end{table}


\section{experiments}

\begin{figure}[h] 
    \centering 
    \includegraphics[width = 0.435 \textwidth]{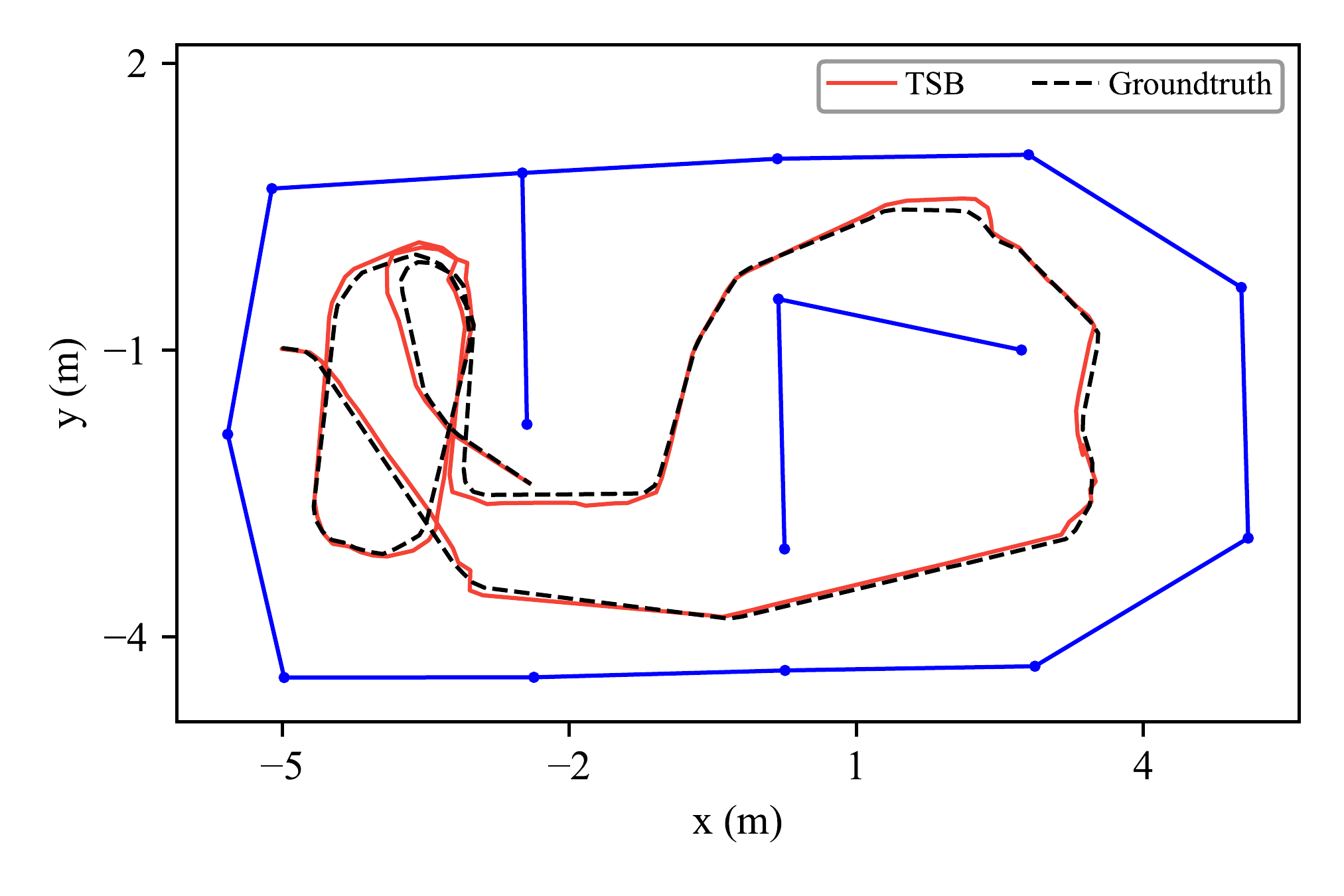} 
    \vspace{-0.3cm}
    \caption{Ground truth and estimated trajectories of Robot 3 during the initial 300 seconds on UTIAS Subset 9. The blue lines represent obstacles obstructing the view.} \label{fig:traj3} 
    \vspace{-0.2cm}
\end{figure}

The experiments were conducted on the publicly available UTIAS dataset \cite{UTIASMultirobotCooperative}. This dataset comprises nine subsets, each lasting from 15 to 70 minutes. Each subset records the odometry, measurements, and ground truth pose of five robots. In addition to inter-robot measurements, the measurements also include relative bearing and distance with respect to known position landmarks. Subset 9 poses a challenge due to obstacles obstructing the view, as illustrated in Figure \ref{fig:traj3}.
In addition to Central and OSB, we compare the proposed method TSB with the following noncentral-equivalent DCL algorithms:
\begin{itemize}
    \item BDA:  employ a block-diagonal approximation to update covariance matrices \cite{luftRecursiveDecentralizedLocalization2018}.
    \item DMV:  be capable of processing various types of robot-to-robot relative measurements \cite{zhuCooperativeLocalizationLimited2019}.
    \item Naive:  disregard cross-correlations among robot pose estimations.
\end{itemize}

These algorithms are evaluated using relative bearing and range measurements as relative position measurements, with odometry utilized as the motion input. Five percent of landmark measurements are utilized to mitigate trajectory drift, while all landmark measurements in Subset 9 are used. Among these algorithms, Central is regarded with the best performance in terms of accuracy and consistency \cite{huangObservabilitybasedConsistentEKF2011}. Thus, Central was tested with over 100 different noise parameter sets for each subset, and the one with the best position accuracy was selected as the noise parameter for that subset.

\begin{figure*}
    \centering
    \includegraphics[width = 0.99 \textwidth]{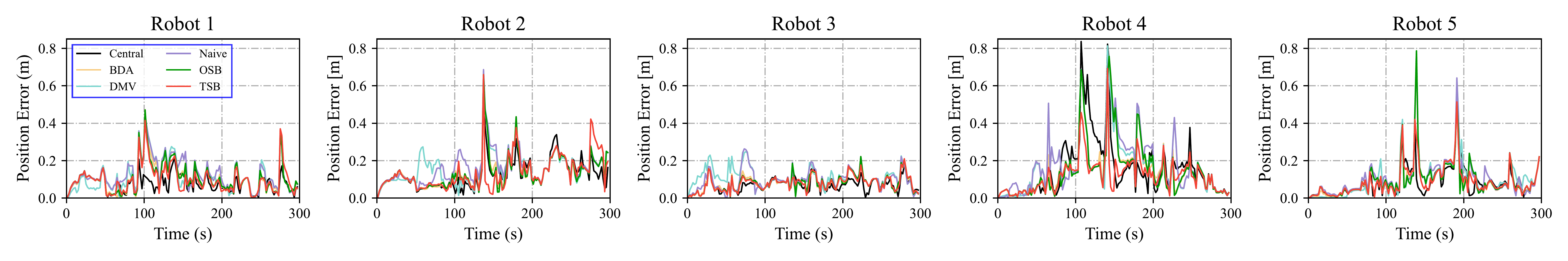}
    \vspace{-0.3cm}
    \caption{5 robot position estimation errors during the first 300s on UTIAS Subset 9}
    \label{fig:t2}
    \vspace{-0.2cm}
\end{figure*}%

Figure \ref{fig:t2} illustrates the estimation errors of the positions of 5 robots during the initial 300 seconds of UTIAS Subset 9. Table \ref{table:utias} reports the RMSE values of the nine subsets. It is evident that the proposed algorithm, TSB, outperforms the other four DCL algorithms in terms of accuracy.
\begin{table}[htbp]
    \centering
    \scriptsize
    \tabcolsep=0.12cm
    \caption{Orientation (deg) and Position (m) RMSE on UTIAS Dataset}
    \label{table:utias}
    \begin{tabular}{cccccccc}
        \toprule
        \footnotesize Set & \footnotesize Central & \footnotesize BDA & \footnotesize DMV    & \footnotesize Naive & \footnotesize OSB             & \footnotesize TSB             \\
        \midrule
        1                 & 16.6 / 0.20           & 18.1 / 0.31       & \textbf{15.9} / 0.39 & 26.2 / 1.11         & 19.2 / 0.30                   & 17.4 / \textbf{0.24}          \\
        2                 & 9.28 / 0.14           & 9.66 / 0.17       & 12.2 / 0.29          & 11.1 / 0.26         & 9.62 / 0.16                   & \textbf{9.55} / \textbf{0.14} \\
        3                 & 8.81 / 0.13           & 9.27 / 0.16       & 9.50 / 0.20          & 9.29 / 0.18         & \textbf{9.20} / 0.15          & 10.2 / \textbf{0.13}          \\
        4                 & 11.2 / 0.12           & 11.9 / 0.19       & 13.2 / 0.27          & 12.7 / 0.28         & 11.6 / 0.18                   & \textbf{11.5} / \textbf{0.15} \\
        5                 & 7.98 / 0.18           & 8.31 / 0.21       & 10.1 / 0.29          & 10.2 / 0.39         & \textbf{8.14} / \textbf{0.19} & 8.23 / \textbf{0.19}          \\
        6                 & 6.83 / 0.12           & 7.23 / 0.14       & 8.93 / 0.20          & 10.5 / 0.22         & 7.00 / 0.13                   & \textbf{6.93} / \textbf{0.12} \\
        7                 & 7.76 / 0.14           & 8.24 / 0.16       & 8.99 / 0.25          & 13.3 / 0.58         & 7.96 / 0.15                   & \textbf{7.95} / \textbf{0.14} \\
        8                 & 12.9 / 0.16           & 18.3 / 0.32       & 27.7 / 0.82          & 17.9 / 0.46         & 17.1 / 0.27                   & \textbf{13.7} / \textbf{0.20} \\
        9                 & 16.9 / 0.24           & 19.1 / 0.31       & 19.4 / 0.31          & 19.3 / 0.30         & 19.1 / 0.30                   & \textbf{17.8} / \textbf{0.25} \\
        \bottomrule
    \end{tabular}
\end{table}

\section{Conclusion }
In this paper, we propose a method to improve the consistency and accuracy of the central-equivalent DCL by performing state estimation based on the transformed error state.
We first design a coordinate transformation for DCL. This transformation
ensures the correct observability property by rendering the propagation Jacobian independent of states and simplifies the propagation of cross-covariance matrices in DCL. Then we integrate the transformation into the server-based framework. Finally, the simulations and the real-world data experiments show that the proposed algorithm outperforms other DCL algorithms in terms of consistency and accuracy.

\bibliographystyle{IEEEtran}
\bibliography{ref}
\end{document}